\documentclass[10pt,twocolumn,letterpaper]{article}

\usepackage{iccv}              

\usepackage{tabularx}
\newcommand{\tname}{MaTe}

\definecolor{iccvblue}{rgb}{0.21,0.49,0.74}
\usepackage[pagebackref,breaklinks,colorlinks,allcolors=iccvblue]{hyperref}

\usepackage{multirow}
\usepackage{booktabs}
\usepackage{multirow}
\usepackage{makecell}
\usepackage{cuted}
\usepackage{algorithm}
\usepackage{algorithmic}
\usepackage{booktabs}
\usepackage{colortbl}
\definecolor{LightBlue}{rgb}{0.9,0.94,1}

\usepackage[accsupp]{axessibility}

\def\paperID{3853} 
\def\confName{ICCV}
\def\confYear{2025}
\usepackage[accsupp]{axessibility}

\title{MaTe: Images Are All You Need for Material Transfer via Diffusion Transformer}

\author{Nisha~Huang$^{1,2}$, Henglin~Liu$^{1}$, Yizhou~Lin$^1$, Kaer~Huang$^3$, Chubin~Chen$^1$, \\Jie~Guo$^{2,\dag}$, Tong-Yee~Lee$^{4}$, Xiu~Li$^{1,\dag}$\\
$^1$Tsinghua University, $^2$PengCheng Laboratory, $^3$Lenovo Research, $^4$National Cheng-Kung University\\
{\tt\small $^1$\{hns24, liu-hl24, yz-lin24, chencb24\}@mails.tsinghua.edu.cn, li.xiu@sz.tsinghua.edu.cn}\\
{\tt\small $^2$\{huangnsh, guoj01\}@pcl.ac.cn $^3$huangke1@lenovo.com $^4$tonylee@mail.ncku.edu.tw}\\}

\begin{document}


\twocolumn[{%
\renewcommand\twocolumn[1][]{#1}%
\maketitle
\begin{center}
    \captionsetup{type=figure}
    \vskip -8.5mm
    \includegraphics[width=0.9\linewidth]{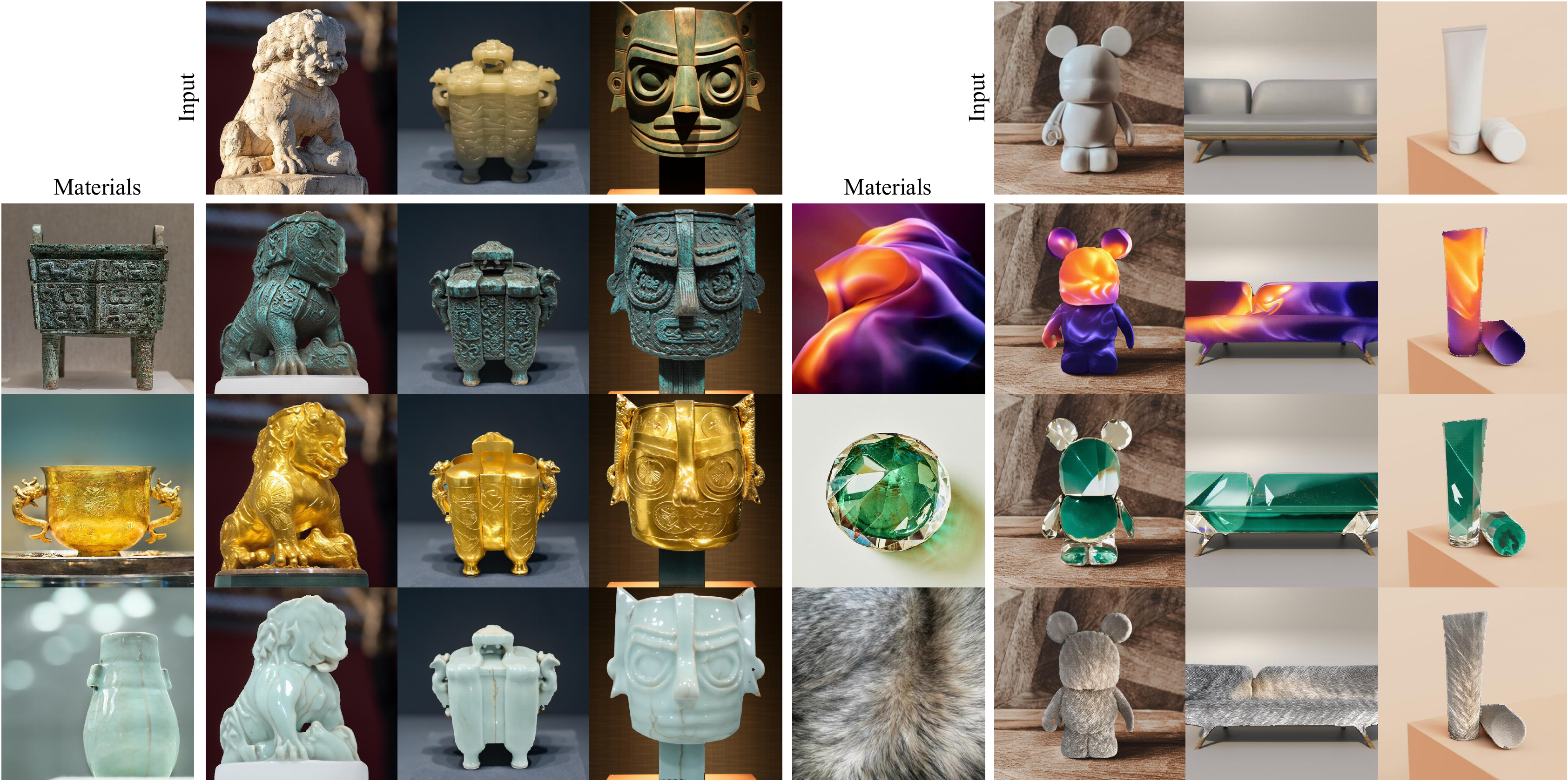}
    \captionof{figure}{
    \textbf{MaTe} is a material transfer method that enables the transformation of textures from a single real-world image without any prior knowledge. This approach is not only capable of successfully extracting texture information from antiques with thousands of years of history but also handles popular computer graphics images, jewelry, and fur materials, providing strong support for design work.
    }
    \label{fig:teaser}
    \end{center}%
}]

\begin{abstract}
Recent diffusion-based methods for material transfer rely on image fine-tuning or complex architectures with assistive networks, but face challenges including text dependency, extra computational costs, and feature misalignment. To address these limitations, we propose MaTe, a streamlined diffusion framework that eliminates textual guidance and reference networks. MaTe integrates input images at the token level, enabling unified processing via multi-modal attention in a shared latent space. This design removes the need for additional adapters, ControlNet, inversion sampling, or model fine-tuning. Extensive experiments demonstrate that MaTe achieves high-quality material generation under a zero-shot, training-free paradigm. It outperforms state-of-the-art methods in both visual quality and efficiency while preserving precise detail alignment, significantly simplifying inference prerequisites.
\end{abstract}    
\section{Introduction}
\label{sec:intro}
Material transfer is a technique that precisely maps the properties of a specific material sample onto the surface of a target object, as shown in Fig.~\ref{fig:teaser}. 
Due to its broad application prospects in digital content creation and industrial design, this technology has garnered attention in recent years.
Traditional material transfer frameworks primarily rely on parametric modeling paradigms, such as optical reflection models based on bidirectional scattering distribution function (BRDF)~\cite{deschaintre2018single,hu2022inverse,henzler2021generative} or procedural texture generation algorithms~\cite{tony,hu2019novel}. These frameworks limit the diversity of the generated results due to the limited size of the basic material library, failing to meet the customization needed for non-uniform composite materials in digital artistic creation.

\begin{figure}
\centering
\includegraphics[width= \linewidth]{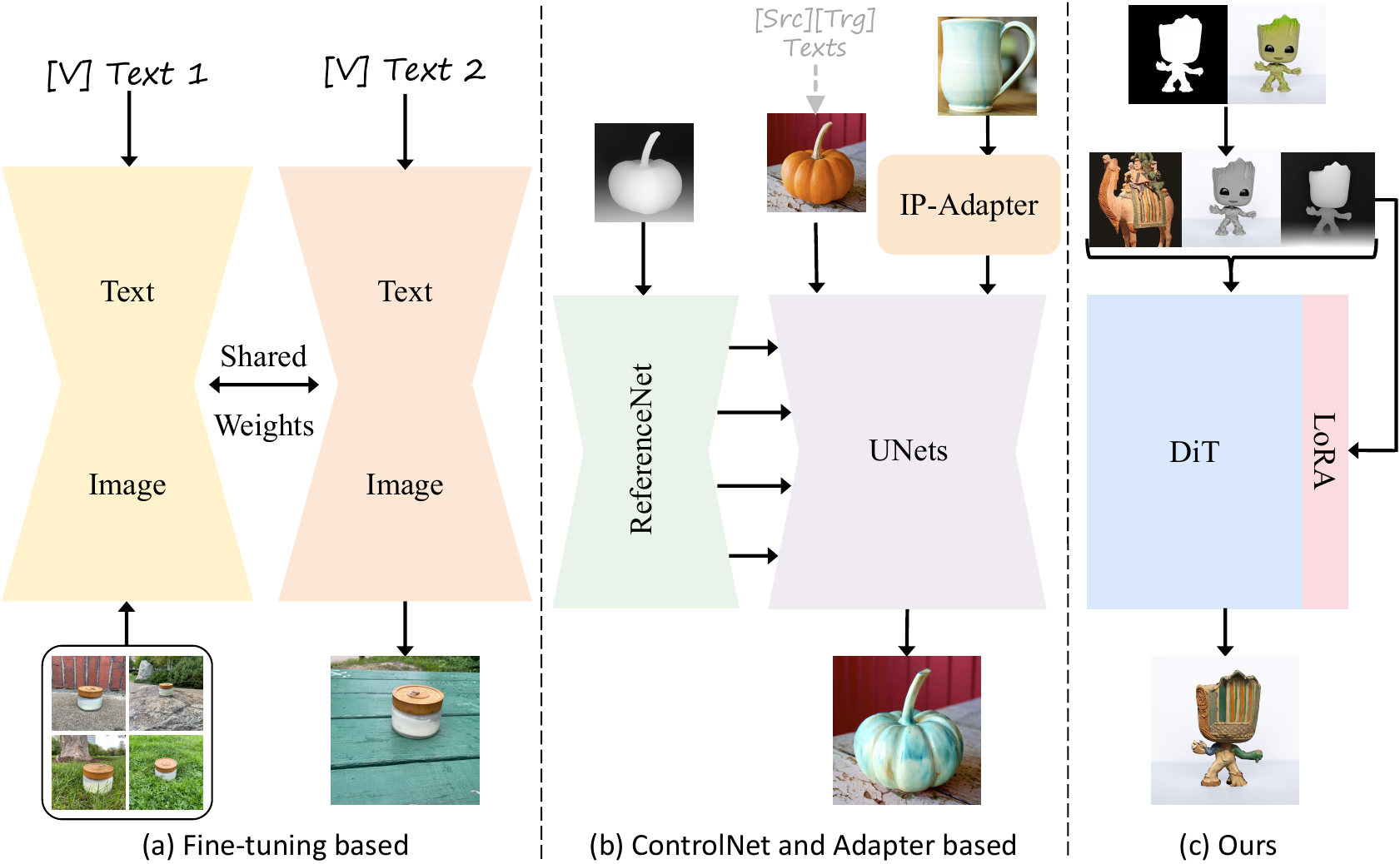}
\vspace{-5mm}
\caption{Simplified structure comparison of different kinds of material transfer methods. Our approach neither relies on fine-tuning image sets/individual images nor requires additional image encoding by IP-Adapters. It only requires basic image information as input, rather than complex text guidance, to obtain high-quality material transfer results.}
\label{fig:insight}
\vspace{-10pt}
\end{figure}

In recent years, inspired by the breakthroughs of diffusion models in conditional generation tasks, diffusion-based material transfer methods~\cite{sharma2024alchemist, wu2024u, lopes2024material} have achieved remarkable improvements in high-quality material transfer effects. Current state-of-the-art (SOTA) approaches often employ fine-tuning of diffusion models with sample sets bound to text identifiers~\cite{ruiz2023dreambooth,zhang2023prospect,wu2024u} (in Fig.~\ref{fig:insight} (a)), implicitly encoding material features through conceptual semantic binding. Although these methods address some shortcomings of traditional pipelines, their heavy reliance on text prompts restricts fine-grained control of material properties. Moreover, the full-parameter fine-tuning paradigm significantly increases training costs and the risk of overfitting. 
The latest studies~\cite{garifullin2025materialfusion, cheng2024zest} have also introduced pre-trained general image encoders IP-Adapters~\cite{ye2023ip-adapter} to extract material features, as well as ControlNet~\cite{zhang2023adding} to inject depth information, as shown in Fig.~\ref{fig:insight} (b). 
However, these methods frequently induce hierarchical decoupling between material and structural information (instead of seamless fusion) during generation, while also suffering from prolonged inference times.

To address the limitations of prior works, we revisit the necessity of employing additional image encoders in material transfer tasks, aligning with cutting-edge trends in image generation research~\cite{esser2024scaling,qwen2vl-flux,catvton,lin2024ctrlx}. The core objective of material transfer lies in ensuring consistent alignment of texture attributes on the target object's surface with the visual details of the material exemplar. When multiple independent conditional control signals are injected in parallel without cross-modal interaction mechanisms, they often induce inter-modal decoupling phenomena. This manifests as isolated encoding of material features, geometric structures, and illumination information in the latent space, where each modality independently influences the generation process rather than enabling feature fusion through collaborative optimization. Such limitations typically result in artifacts like material-geometry misalignment, where material textures fail to conform to surface curvature, and illumination-reflection inconsistency, where specular highlights deviate from light source directions.

The above insights, combined with the emergence of the Diffusion Transformer (DiT) paradigm~\cite{peebles2023scalable}, motivate us to re-examine fundamental approaches for material transfer. 
Although IP-Adapter and ControlNet effectively represent material and structural information individually, limitations such as inconsistent feature spaces, lack of interaction modules, and information competition during generation hinder the natural fusion of material, depth, and input images in diffusion models (in Fig.~\ref{fig:comparison} (d)-(g)).
We consider that semantic alignment across multimodal features and their interactive potential within a shared latent space should be prioritized, thus projecting all image conditions into a unified latent representation.
We design a novel MaTe architecture that processes image conditions in a unified manner through multimodal attention, eliminating reliance on complex control modules. The Unified-Sequence Processing and Cross-Bias Modulation mechanisms in MaTe enable the denoising network to jointly handle all image signals, requiring only lightweight Low-Rank Adaptation (LoRA)~\cite{hu2022lora} to augment depth information.

In summary, we highlight our contributions as follows:

\begin{itemize}
    \item We propose MaTe, a concise material transfer architecture that enables semantic interaction between image conditions alongside zero-shot control, significantly reducing architectural complexity and inference time.
    \item We simplify the inference process, eliminating the need for fine-tuning or manually setting text information. We achieve the inference of all necessary information from input images using existing pre-trained diffusion models.
    \item We construct a diverse and extensive real-world material transfer evaluation dataset. Our method produces high-quality material transfer results with consistent details, outperforming SOTA baselines in both qualitative and quantitative analyses.
\end{itemize}

\section{Related work}
\label{sec:relatedwork}
\subsection{Image-guided generation.}
Image-guided generation is a challenging yet promising field, leveraging visual guidance for content creation. 
The emergence of diffusion models~\cite{rombach2022high,peebles2023scalable,huang2022draw,huang2024diffstyler} has significantly propelled advancements in this field, enabling SOTA performance across a wide range of generative visual tasks, such as image-to-image translation~\cite{ye2023ip-adapter,huang2025creativesynth,huang2025artcrafter}, subject-driven image generation~\cite{ruiz2023dreambooth,hertz2022prompt}, etc.
DreamBooth~\cite{ruiz2023dreambooth} and Textual Inversion~\cite{galimage} leverage transfer learning for text-to-image (T2I) diffusion models to enable customized concept generation via full parameter fine-tuning or word vector optimization.
To enhance the controllability of image-guided generation, adapter-based architectures have emerged as bridges between external control signals (e.g., sketches) and diffusion models.
ControlNet~\cite{zhang2023adding} first validates an adapter that can be trained to capture task-specific input conditions, while T2I-adapter~\cite{mou2024t2i} employs a lightweight adapter to achieve fine-grained control in the color and structure of the generated images.
IP-Adapter~\cite{ye2023ip-adapter} allows for more flexible and intuitive control of the generation process, expanding the capabilities of image-guided generation. 
These methods demonstrate image-conditioned generation strategies with varying specificity, suggesting potential pathways for addressing material transfer challenges.

\subsection{Material acquisition and transfer.}
Material acquisition and transfer represent a realm of research considering illumination conditions, object geometry, and physical properties of materials. 
Traditional 3D material transfer methods, such as Text2tex~\cite{chen2023text2tex}, TEXTure~\cite{richardson2023texture}, and TextureDreamer~\cite{yeh2024texturedreamer}, rely on 3D geometric shapes and lighting estimation, followed by careful adjustment of material properties.
As a result, the quality and diversity are restricted, presenting limited and unsatisfactory results.
In contrast, 2D-to-2D material transfer, bypassing ground-truth 3D data, is challenging yet highly practical.
Prevalent approaches based on fine-tuning including DreamBooth~\cite{ruiz2023dreambooth}, Material Palette~\cite{lopes2024material}, Prospect~\cite{zhang2023prospect} and U-VAP~\cite{wu2024u} fine-tune diffusion models on small sample sets associated with text identifiers. 
While existing methods mitigate conventional pipeline limitations, text prompts reliance restricts control and fine-tuning risks computational costs, and overfitting. 
Another category of methods, such as MaterialFusion~\cite{garifullin2025materialfusion} and ZeST~\cite{cheng2024zest}, uses extra encoder modules like IP-Adapter~\cite{ye2023ip-adapter} or ControlNet~\cite{zhang2023adding} to extract material features.
When multiple image conditions are injected into the network in parallel, the material and structure information in the results are separated into two independent layers, lacking semantic integration (see Fig.~\ref{fig:comparison} (d)-(g)). The latest SOTA method MaterialFusion~\cite{garifullin2025materialfusion} applies DDIM inversion~\cite{songdenoising} to both the material and the input images, which increases the sampling steps and time by more than double.
Different from methods that rely on text-assisted fine-tuning or complex architectures with additional reference networks, our method MaTe projects material, content, and depth images into the same latent space and performs semantically aligned generation.

\section{Method}
\begin{figure*}
\centering
\includegraphics[width= 0.95\linewidth]{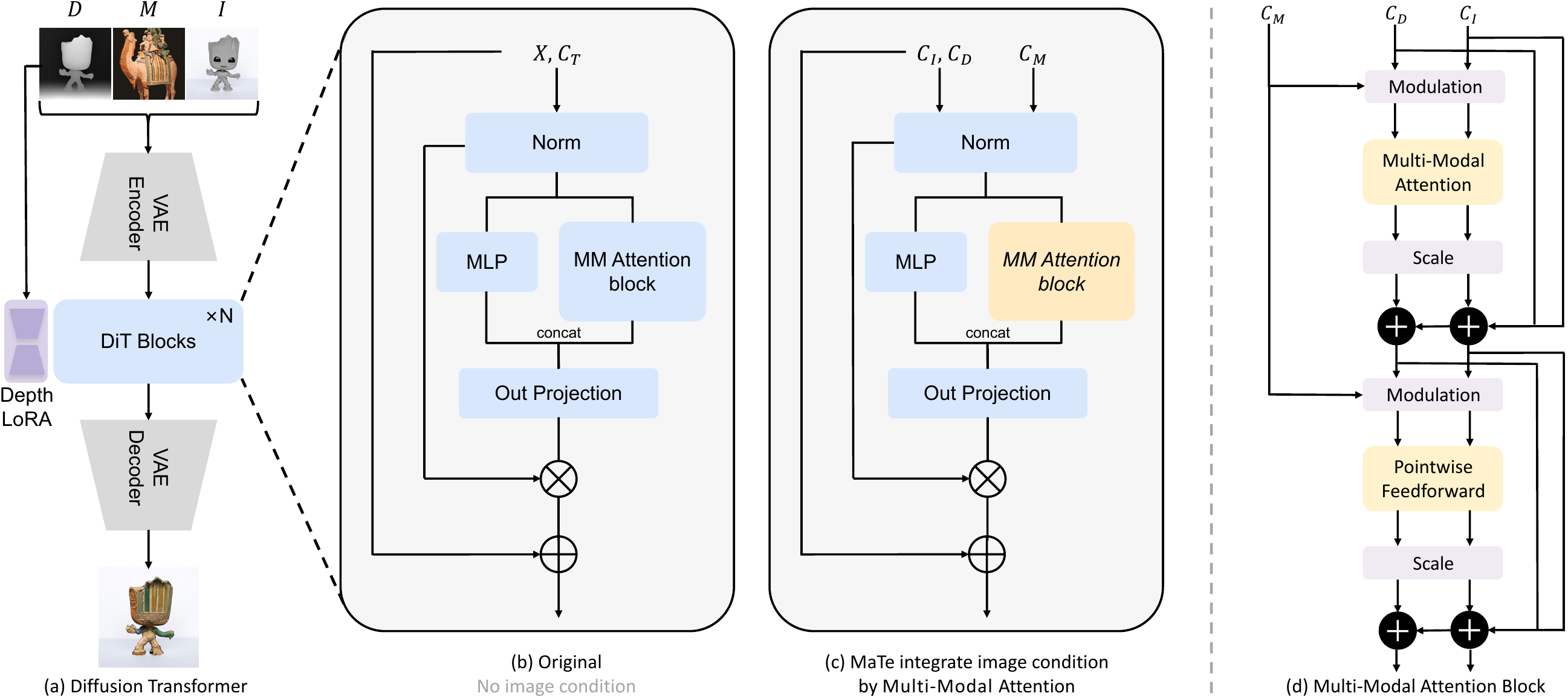}
\vspace{-2mm}
\caption{Our method achieves high-quality material transfer by simply passing three types of image tokens (material image tokens $C_M$,  depth image tokens $C_D$, and illumination image tokens $C_I$) through the multimodal attention mechanism of DiT for interaction, ensuring that they remain in the same feature space throughout the diffusion process. It eliminates the need for unnecessary image fine-tuning and text guidance, resulting in a streamlined inference process.}
\label{fig:pipeline}
\vspace{-10pt}
\end{figure*}

MaTe is a training-free, non-text prompts, and zero-shot image-to-image framework with structure and material control. Given the target image \( I_{\text{input}} \) and the material image \( I_M \), MaTe produces an output image \( I_o \) that inherits the structure from \( I_{\text{input}} \) and the material from \( I_M \).
Our approach is illustrated in Fig.~\ref{fig:pipeline} and summarized as follows:
After providing the target image \( I_{\text{input}} \) and the material image \( I_M \), in Sec.~\ref{sec:3.3}, we obtain the illumination image \( I \) and the depth image \( I_D \). \( I_M \), \( I \), and \( I_D \) undergo \textit{Unified-Sequence Processing} and \textit{Cross-Bias Modulation} through our designed MaTe architecture to achieve control over structure and material. Additionally, LoRA is used to enhance the depth effect, and \textit{Background-Preserving Blending} is employed to strengthen the consistency of the foreground and background through Sec.~\ref{sec:3.3}.

\subsection{Preliminary}
\noindent\textbf{Rectified-Flow Models.}
Generative models aim to define a mapping from samples \( x_1 \) from a noise distribution \( p_1 \) to samples \( x_0 \) from a data distribution \( p_0 \), where \( p_0 \) represents real images in image generation tasks. Rectified flows \cite{lipmanflow, liuflow} define a forward process that constructs paths between distributions \( p_0 \) and \( p_1 \) as straight trajectories, as shown in Eq. \ref{eqn:flow_path}, where \( p_1 = \mathcal{N}(0, 1) \). Here, the forward process is time-dependent due to timestep \( t \).
\begin{equation}
    \label{eqn:flow_path}
    x_t = (1 - t)x_0 + t\epsilon, \quad \epsilon \sim N(0, 1)
\end{equation}
\noindent To learn this mapping, a network is trained with parameters $\theta$, to estimate the velocity \( v \) of the rectified flow, represented by \( v_\theta \). By adopting the reparameterization from \cite{esser2024scaling}, this velocity prediction network can serve as a noise prediction network, \( \epsilon_\theta \), optimized using the Conditional Flow Matching (CFM) objective formulated as Eq. \ref{eqn:cfm_objective}.
\begin{equation}
    \label{eqn:cfm_objective}
    \mathcal{L}_{CFM} = -\frac{1}{2} \mathbb{E}_{t \sim \mathcal{U}(t), \epsilon \sim \mathcal{N}(0, I)}[w_t \lambda_t'||\epsilon_{\theta}(x_t, t) - \epsilon||^2]
\end{equation}
\noindent Here, \( \lambda_t' \) represents the re-parametrized signal-to-noise ratio, and \( w_t \) is a time-dependent weighting function.

\noindent\textbf{Multi-Modal Diffusion Transformers.}
The DiT model~\cite{peebles2023scalable}, is employed in architectures like FLUX.1~\cite{flux2024}, Stable Diffusion 3~\cite{esser2024scaling}, and PixArt~\cite{chen2023pixart} use transformer as denoising network to refine noise image tokens iteratively.
As shown in Fig.~\ref{fig:pipeline} (a)), the DiT model processes two token types: noise image tokens ${X}\in \mathbb{R}^{N
\times d}$ and text condition tokens ${C}_{\text{T}}\in \mathbb{R}^{M \times d}$, where $d$ is the embedding dimension, $N$ and $M$ are the number of image and text tokens respectively. Throughout the network, these tokens maintain consistent shapes as they pass through multiple transformer blocks.
In FLUX.1, each DiT block consists of layer normalization followed by multi-modal attention (MMA)~\cite{pan2020multi}, which incorporates rotary position embedding (RoPE)~\cite{su2024roformer}
to encode spatial information.

The multi-modal attention mechanism then projects the position-encoded tokens into query $Q$, key $K$, and value $V$ representations. It enables the computation of attention between all tokens:
\begin{equation}
  \text{MMA}([{X};{C}_{\text{T}}]) = \text{softmax}\left(\frac{QK^{\top}}{\sqrt{d}}
  \right)V, \label{eq:mma}
\end{equation}
where $[{X};{C}_{\text{T}}]$ denotes the concatenation of image and text tokens. This formulation enables bidirectional attention.
Building on the DiT architecture with FLUX.1 as the implementation basis, we aim to develop MaTe, a framework that balances superior performance with minimalist design, guiding material transfer exclusively via visual conditions.

\subsection{Depth and Illumination Guidance}
\label{sec:3.2}

\noindent\textbf{Lighting Consistency.}
Diffusion models typically start sampling from random noise, which can lead to the lighting loss of optical priors in the initial image.
As ZeST~\cite{cheng2024zest}, we adopt a foreground grayscale image \( I \) to retain illumination without base color prior.
The denoising process is initialized using the following formula:
\begin{equation}  
I = F \odot I_{gray} + (1 - F) \odot I_{input}  
\label{grayscale}  
\end{equation}  
Here, \( F \) denotes the object mask, and \( I_{gray} \) represents the grayscale image. This formula achieves the decoupling of lighting preservation and material interference: The component \( (1 - F) \odot I_{input} \) retains the environmental lighting characteristics (direction, intensity, and hue), while \( F \odot I_{gray} \) eliminates the inherent base color contamination of the source object while maintaining geometric shadow information. 
As shown in the ablation study in Fig.~\ref{fig:ablation_light}, the illumination image $I$ can better preserve the illumination information compared to random noise.

\noindent\textbf{Geometric Guidance Optimization.}
Given the inherent challenges in decoupling geometric and material attributes within images and the requirement for additional training data, we propose an enhanced solution that integrates depth image $D$ as geometric priors within the diffusion model, reinforcing the structural representation of target objects through a lightweight standard way. Comparative experiments (detailed in Sec.~\ref{ablation}) demonstrate our rejection of conventional ControlNet architectures (e.g., \cite{fluxdepth2} with 1.59B parameters) in favor of a FLUX.1 depth-aware LoRA module~\cite{hu2022lora} (620M parameters). Experimental validation confirms this design achieves a $1.8×$ speedup in inference for material transfer tasks while maintaining geometric fidelity.

\subsection{MaTe}
\label{sec:3.3}


\noindent\textbf{Architecture Designing.}
To achieve both versatility architectural changes, {\tname} first reuses the VAE encoder~\cite{Kingma2013AutoEncodingVB, rombach2022high} from the base DiT, treating the illumination image $I$ as the noise image and projecting the material image into the same latent space as the noise image tokens. This approach contrasts sharply with previous methods that rely on separate feature extractors (e.g., IP-Adapter~\cite{ye2023ip-adapter,cheng2024zest,garifullin2025materialfusion}), significantly reducing architectural complexity.
We achieve material transfer by leveraging the spatial correspondence in the multi-modal attention block, without the need for targeting specific subjects.
The encoded depth condition tokens ${C}_{\text{D}}$ share the same dimensionality and latent space as the illumination image tokens ${C}_{\text{I}}$, enabling them to be directly processed by the Transformer blocks. Since both condition and image tokens reside in the same latent space, {\tname} leverages the existing DiT blocks to jointly process them. 
Notably, we innovatively propose \textit{Cross-Bias Modulation}, a novel mechanism that enables controllable cross-modal interaction through structured logarithmic bias, thereby achieving free control among conditions.

\noindent\textbf{Low-Rank Adaptation.}
The LoRA approach enhances parameter efficiency by freezing the pre-trained weight matrices and introducing additional trainable low-rank matrices within the neural network. This method is based on the observation that pre-trained models exhibit low ``intrinsic dimension''. Specifically, for a weight matrix \( W \in \mathbb{R}^{n \times m} \) in the diffusion model \( \epsilon_{\theta} \), incorporating a LoRA module involves updating \( W \) to \( W' \), defined as \( W' = W + BA \). Here, \( B \in \mathbb{R}^{n \times r} \) and \( A \in \mathbb{R}^{r \times m} \) are matrices of a low-rank factor \( r \), satisfying \( r \ll \min(n, m) \).
In our work, we employ a LoRA module to control depth information~\cite{fluxdepth}. By assigning a specific weight \( w \) to the LoRA module, we can modulate its influence on the generation process:
\begin{equation}
W' = W +  w \times B A.
\end{equation}
The weight $w$ is typically a hyperparameter determined through empirical tuning and is used to balance the relationship between depth information and other generative features. This allows LoRA to effectively incorporate depth information into the diffusion model to produce images that conform to the depth structure, unlike previous material transfer methods that use ControlNet.


\begin{figure*}
\centering
\includegraphics[width= 0.95\linewidth]{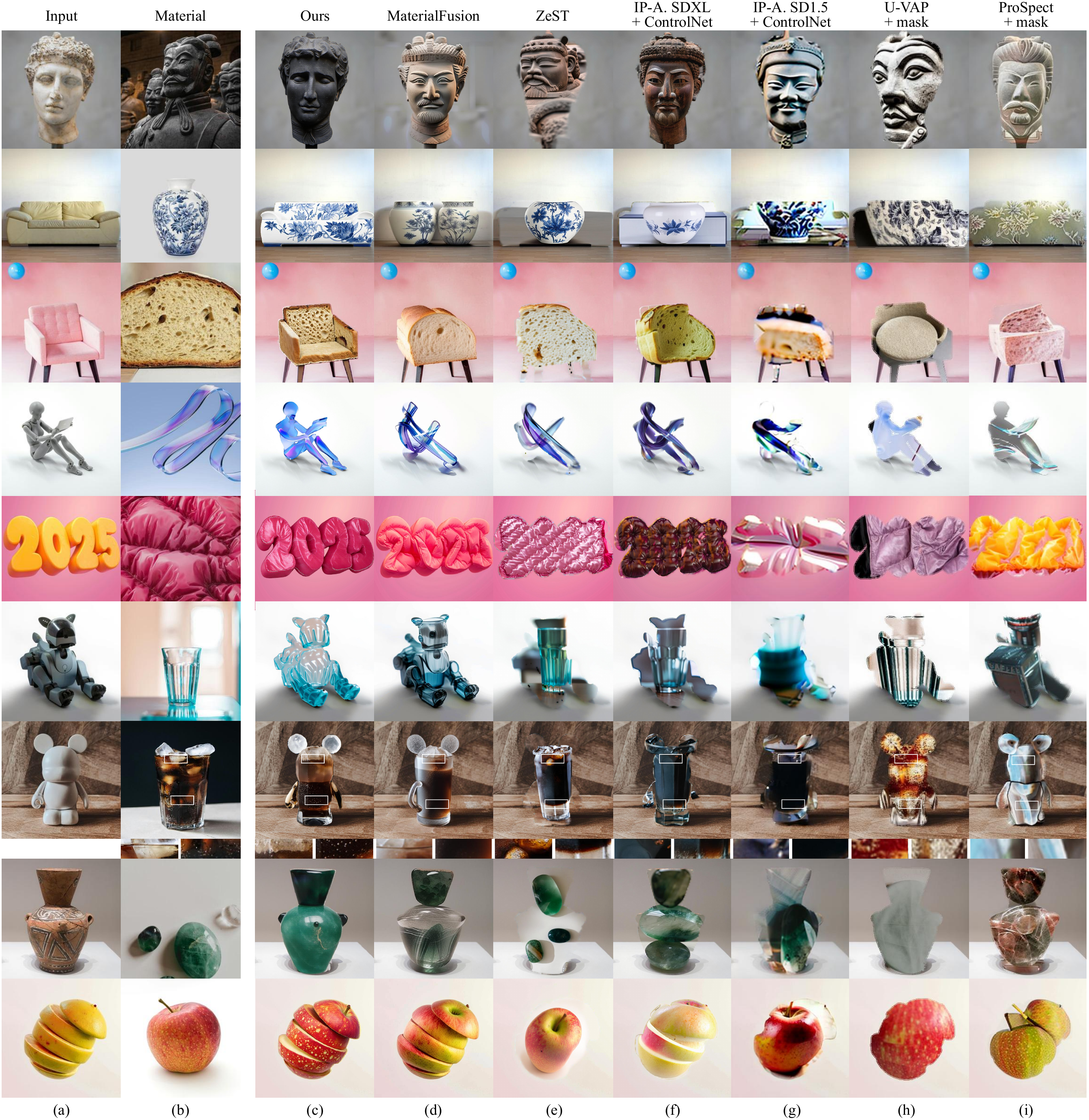}
\vspace{-3mm}
\caption{Qualitative comparison on the MTB dataset. MaTe demonstrates a distinct advantage in handling complex materials. (d)-(g) are based on work utilizing IP-Adapter and ControlNet, while (h) and (i) are based on fine-tuning with input images and texts.
}
\label{fig:comparison}
\vspace{-10pt}
\end{figure*}

\noindent\textbf{Unified-Sequence Processing.}
%
Previous methods, such as ControlNet~\cite{zhang2023adding} and T2I-Adapter~\cite{mou2024t2i}, introduced condition images into the model through direct feature addition:
\begin{equation}
  {X}\leftarrow{X}+{C}_{\text{D}},
\end{equation}
where the condition features ${C}_{\text{D}}$ are spatially aligned and added to the noise image tokens ${X}$. While this method is effective for spatially aligned tasks, it faces two limitations: (1) it lacks flexibility in non-aligned scenarios where spatial correspondence does not exist, and (2) the rigid addition operation constrains potential interactions between condition and image tokens.
In contrast, MaTe directly concatenates condition tokens with the image tokens \([ {C}_{\text{M}}; {C}_{\text{I}}; {C}_{\text{D}} ]\) for multi-modal attention processing. This new operation enables flexible token interactions through DiT's multi-modal attention mechanism, allowing direct relationships to emerge between any pair of tokens without imposing strict spatial constraints.

\noindent\textbf{Cross-Bias Modulation.}
While {\tname}'s unified sequence processing and multi-modal attention enable effective token interactions during training, practical applications often require adjustable conditioning strength at test time.
We achieve this by introducing a bias term into the multi-modal attention computation. Specifically, for a given strength factor $\gamma$, we modify the attention operation in Equation~\ref{eq:mma} to:
\begin{equation}
  \text{MMA}([{C}_{\text{M}};{C}_{\text{I}};{C}_{\text{D}}]) = \text{softmax}\left(\frac{QK^{\top}}{\sqrt{d}}
  + B(\gamma)\right )V,
\end{equation}
where $B(\gamma)$ is a bias matrix modulating the attention between concatenated tokens $[{C}_{\text{M}};{C}_{\text{I}};{C}_{\text{D}}]$. Given ${C}_{\text{M}}\in \mathbb{R}
^{M \times d}$ and ${C}_{\text{I}},{C}_{\text{D}}\in \mathbb{R}^{N \times d}$, the bias matrix has the structure:
\begin{equation}
\setlength{\arraycolsep}{2pt}  
B(\gamma) = \begin{bmatrix}
\mathbf{0}_{\scriptscriptstyle M \times M} & 
\log(\gamma)\mathbf{1}_{\scriptscriptstyle M \times N} & 
\log(\gamma)\mathbf{1}_{\scriptscriptstyle M \times N} \\[2pt]
\log(\gamma)\mathbf{1}_{\scriptscriptstyle N \times M} & 
\mathbf{0}_{\scriptscriptstyle N \times N} & 
\mathbf{0}_{\scriptscriptstyle N \times N} \\[2pt]
\log(\gamma)\mathbf{1}_{\scriptscriptstyle N \times M} & 
\mathbf{0}_{\scriptscriptstyle N \times N} & 
\mathbf{0}_{\scriptscriptstyle N \times N}
\end{bmatrix}.
\end{equation}
The bias matrix structure ensures that the attention patterns within each modality (main diagonal blocks) remain unchanged, while the interaction strength between modalities ($C_M$ and $C_I/C_D$) is modulated by introducing a $\log(\gamma)$ bias term. During the testing phase, when $\gamma \rightarrow 0^+$ (taking an extremely small value), the influence of the condition can be eliminated, and when $\gamma > 1$, the condition guidance is enhanced by increasing the cross-modal attention weights. This method allows for flexible control over the integration strength of image conditions and generation results without the need to retrain the model.

\noindent\textbf{Background-Preserving Blending.}
A simple method for preserving the background is to replace the generated background with the original one, taken from the input image: ${x'} \odot m + x \odot (1 - m)$. The obvious issue is that combining the two images in this way fails to produce a coherent, seamless result.
Our key hypothesis is that at each step of the diffusion process, a noise latent representation is projected onto a manifold of natural images that are noised to a certain level. While blending two noise images (from the same level) may yield a result that likely lies outside the manifold, the next diffusion step projects the result onto the next level manifold, thereby ameliorating the incoherence.

Therefore, at each stage, starting from the latent representation $x_t$, a diffusion step is performed according to the conditional prompt, yielding a latent representation denoted as $x_{\text{gen}}$. This is combined with the noise version of the original input image $x_{\text{in}}$ obtained from the input image. The two latent representations are now blended using the mask:
\begin{equation}
x_{t-1} = x_{t-1,\text{gen}} \odot m + x_{t-1,\text{in}} \odot (1 - m),
\end{equation}
and the process is repeated. In the final step, the entire region outside the mask is replaced with the corresponding region from the input image, thus strictly preserving the background. This straightforward but effective method for implementing foreground-background fusion is free from training diffusion architectures based on Inpainting.

\section{Experiments}
\subsection{Datasets}
To comprehensively evaluate the performance of various material transfer methods, we have established a freely available open-source dataset Material Transfer Benchmark (MTB) specifically designed to assess the efficacy of material transfer techniques. This dataset comprises $60$ material images and $30$ photographs of target objects as inputs. All photos are from Unsplash~\cite{unsplash}, a website known for its copyright-free content, ensuring the dataset's accessibility and usability for research purposes. Examples of the dataset are provided in the Supplementary Materials for reference.

\begin{table*}[h!]
    \centering
    \resizebox{\textwidth}{!}{
    \begin{tabular}{l|ccc|ccc|ccc|ccc}
        \toprule
        & \multicolumn{3}{c|}{Generation Quality} & \multicolumn{3}{c|}{User Study} & \multirow{2}{*}{\makecell{Base\\model}} & \multirow{2}{*}{\makecell{Preprocessing\\time (s)}} & \multirow{2}{*}{\makecell{Inference\\time (s)}} & \multirow{2}{*}{\makecell{Fine-\\tuning}} & \multirow{2}{*}{\makecell{Text\\conditions}} & \multirow{2}{*}{\makecell{Reference\\Networks}}\\
        \cline{2-7} 
        & SSIM $\uparrow$ & LPIPS $\downarrow$ & CLIP $\uparrow$ & Texture $\uparrow$ & Structure $\uparrow$ & Overall $\uparrow$ & &  &  & & \\
        \midrule
        ProSpect~\cite{zhang2023prospect}  &0.8048  &0.1693 &0.8009 & 2.57\% & 2.57\% & 1.84\% & SD1.4~\cite{rombach2022high} &200   &16  &\checkmark  &\checkmark &-\\
        U-VAP~\cite{wu2024u} &0.7176  &0.2612 &0.7698 & 0.37\% & 0.18\% & 0.00\% & SD1.5~\cite{rombach2022high}  &522  & 7 &\checkmark  &\checkmark &- \\
        IP-Adapter SD1.5~\cite{ye2023ip-adapter}  &0.7489  &0.2537 &0.8271 & 0.74\% & 0.37\% & 0.00\% & SD1.5~\cite{rombach2022high} & 0 &10  &-  &- &\checkmark\\
       IP-Adapter SDXL~\cite{ye2023ip-adapter}  &0.8152  &0.1876 &0.8358 & 7.35\% & 2.94\% & 3.49\% & SDXL~\cite{podell2023sdxl}  & 0 &20  &-  &- &\checkmark\\
       \rowcolor[gray]{0.965} ZeST~\cite{cheng2024zest}  &0.7231  &0.2430 &\underline{0.8627} & 14.34\% & 0.18\% & 0.37\% &  SDXL~\cite{podell2023sdxl} & 0  &21 &-  &- &\checkmark\\
        \rowcolor[gray]{0.945} MaterialFusion~\cite{garifullin2025materialfusion}  &\underline{0.8263}  &\underline{0.1565} &0.8521 & 11.21\% & 20.04\% & 10.48\% & SD1.5~\cite{rombach2022high} & 0 &43  &-  &\checkmark &\checkmark\\
        \rowcolor{LightBlue} MaTe(Ours) &\textbf{0.8617}  &\textbf{0.1319} &\textbf{0.8825} & 63.42\% & 73.71\% & 83.82\% &Flux.1~\cite{flux2024}  & 0 &11  & - & -  & -\\
        \bottomrule
    \end{tabular}}
        \vspace{-2mm}
    \caption{Quantitative comparisons with other methods are presented. MaTe and baselines are compared across common quantitative metrics, user studies, and model efficiency. The results that represent the current SOTA baselines for material transfer are highlighted in gray, with the best results shown in bold and the second-best results underlined.}
    \label{tab:comparison}
    \vspace{-4mm}
\end{table*}

\subsection{Implementation Details}
We build our method upon FLUX.1~\cite{flux2024}, a latent rectified flow transformer for image generation. 
For our experiments, we set inference steps to $8$ and configured the image generation size to $1024 \times 1024$ pixels. Additionally, we utilized a classifier-free guidance scale of $30$. All experiments were conducted on a single NVIDIA $A100$ GPU.

For comparison experiments settings, we compare MaTe with MaterialFusion~\cite{garifullin2025materialfusion}, ZeST~\cite{cheng2024zest}, IP-Adapter SDXL~\cite{ye2023ip-adapter}, IP-Adapter SD1.5~\cite{ye2023ip-adapter}, U-VAP~\cite{wu2024u}, and ProSpect~\cite{zhang2023prospect} in the task of material transfer.
To enhance the fairness of the experiments, we added the corresponding version of the ControlNet depth control models~\cite{control_v11f1p_sd15_depth,controlnet-depth-sdxl-1.0} to the pipelines for IP-Adapter SD1.5 and IP-Adapter SDXL. Additionally, we incorporated mask segmentation operations for IP-Adapters, U-VAP, and ProSpect.

\begin{figure}
\centering
\includegraphics[width= \linewidth]{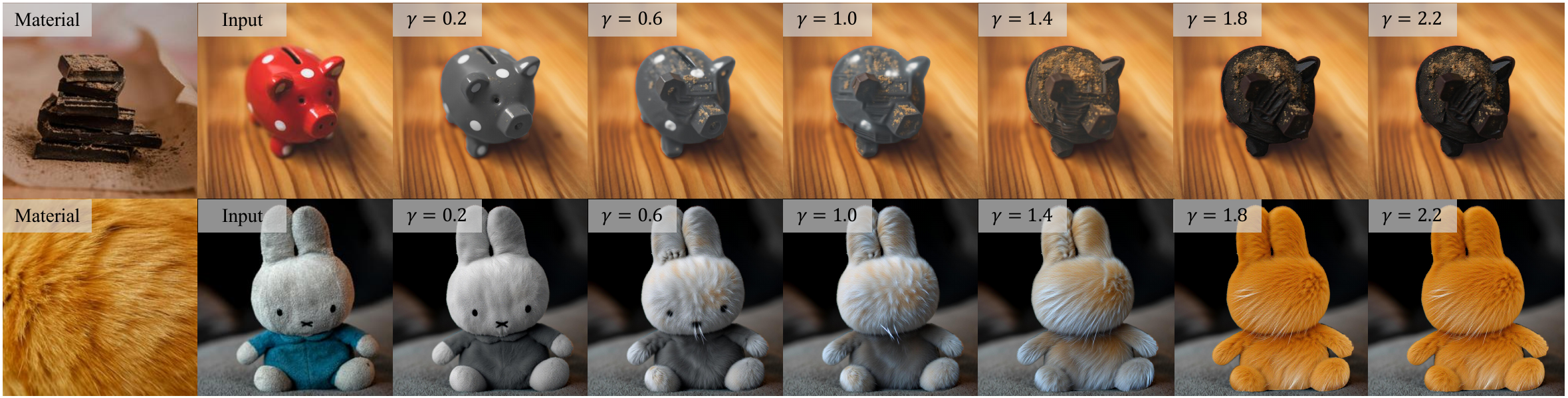}
\vspace{-6mm}
\caption{Material image effect intensity ablation experiment.
}
\label{fig:ablation_strength}
\vspace{-8pt}
\end{figure}
\begin{figure}
\centering
\includegraphics[width= \linewidth]{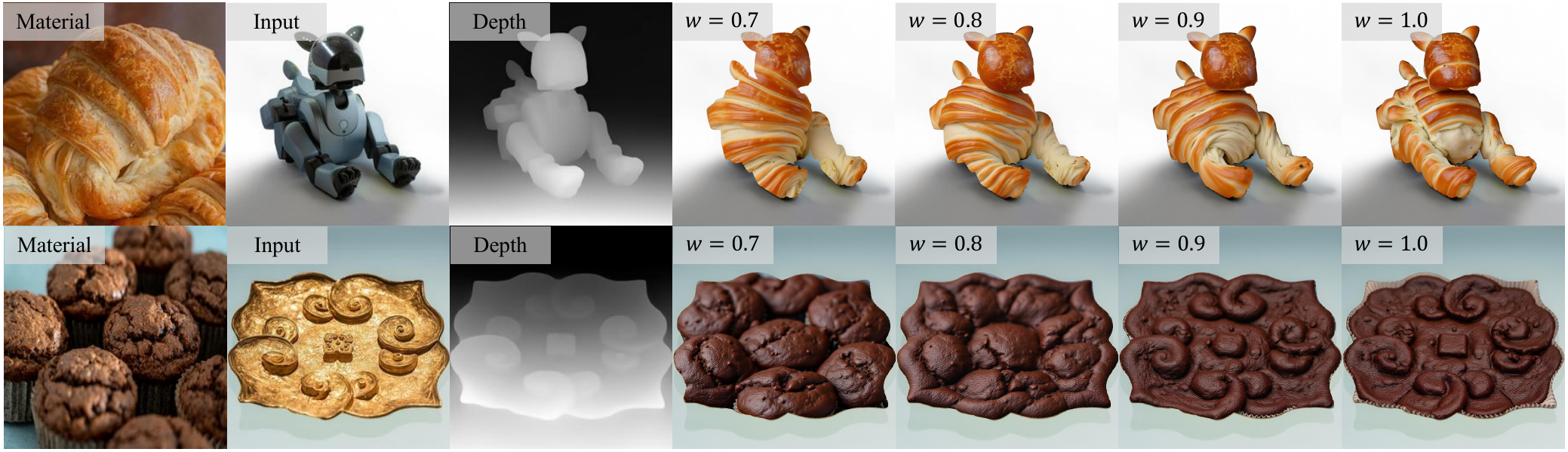}
\vspace{-5mm}
\caption{Ablation experiment on the depth control parameter $w$.
}
\label{fig:ablation_lora_depth}
\vspace{-10pt}
\end{figure}

\subsection{Qualitative Comparison}
As shown in Fig.~\ref{fig:comparison}, our method faithfully retains the structure of the input images. Material transfer based solely on image conditions is not possible with previous methods. MaTe ingeniously transfers appearance from the material image, capturing the correspondence between the subject and the material, achieving a balanced alignment of structural preservation and material representation. In contrast, personalized methods based on image fine-tuning~\cite{zhang2023prospect,wu2024u} did not show significant material transfer effects. Additionally, methods based on ControlNet + IP-Adapter~\cite{cheng2024zest,garifullin2025materialfusion,zhang2023adding,ye2023ip-adapter} (Fig.~\ref{fig:comparison} (d)-(g)) often fail to maintain structure or transfer the appearance of the material. The latest state-of-the-art method, MaterialFusion, also struggles to balance the expression of structure and material information.

Specifically, in the chair case, we accurately captured the hollow structural features of the chair, which other methods failed to retain. In the composite material ribbon example, we restored material properties such as color gradient, transparency, and gloss. Furthermore, in the cola ice cube case, MaTe maintained the shape and spatial layout, which other methods failed to preserve. In the apple example, although the material image did not show the inner flesh, our method still accurately presented the apple's material characteristics, to some extent indicating that the model aligned the material information and expressed a certain level of prior understanding, while other methods had deviations in texture details, demonstrating the accuracy of our method in material understanding and transfer.

\subsection{Quantitative Comparison}
\noindent\textbf{Effect Comparison.}
We conduct quantitative evaluations and comparisons with state-of-the-art material transfer methods on the MTB dataset. We use the Structural Similarity Index (SSIM)~\cite{wang2004image} to measure the structural consistency between the input images and the results, the Learned Perceptual Image Patch Similarity (LPIPS)~\cite{zhang2018unreasonable} to assess the content detail consistency between the input images and the results, and the Contrastive Language–Image Pre-training (CLIP)~\cite{radford2021learning} to evaluate the similarity between the materials and the results.
The results are shown in Table~\ref{tab:comparison}. Our method outperforms all other methods across all metrics, demonstrating the effectiveness of our model architecture for the material transfer task.

\noindent\textbf{User Study.}
The human subjective evaluation is divided into three parts: texture, structure, and overall results. We asked users to select their favorite outcome from all methods in each aspect. We sought a total of $ 16$ evaluators who have experience in the vision and graphics field. Based on $34$ sets of comparative results (each set includes $7$ methods), evaluations were conducted from three perspectives, yielding $1,632$ voting results. The percentages in columns $5-7$ of Table~\ref{tab:comparison} reflect the voting situation of the users, indicating that the results generated by our proposed MaTe are more favored in all three aspects.
We note the difference between objective and subjective evaluation metrics: while objective metrics assess aspects in isolation, users may integrate information across aspects despite being provided with separate options.
Human preference implies that our generated outcomes have struck a better balance among texture, structure, and overall visual appeal.
For detailed information, please refer to the Supplementary Materials.

\noindent\textbf{Efficiency comparison.}
Column 8 of Table~\ref{tab:comparison} lists the diffusion model versions used by different methods, among which we use Flux.1 as the basic model, highlighting the innovation and advancement of MaTe.
Columns 9-10 further compare preprocessing time and inference time. By eliminating ControlNet or additional text or image encoders, our framework achieves significant time reduction throughout the inference process.
Finally, columns 11-13 illustrate MaTe is the first method that does not require fine-tuning, text, or reference networks, demonstrating its user-friendliness and practicality.

\begin{figure}
\centering
\includegraphics[width= \linewidth]{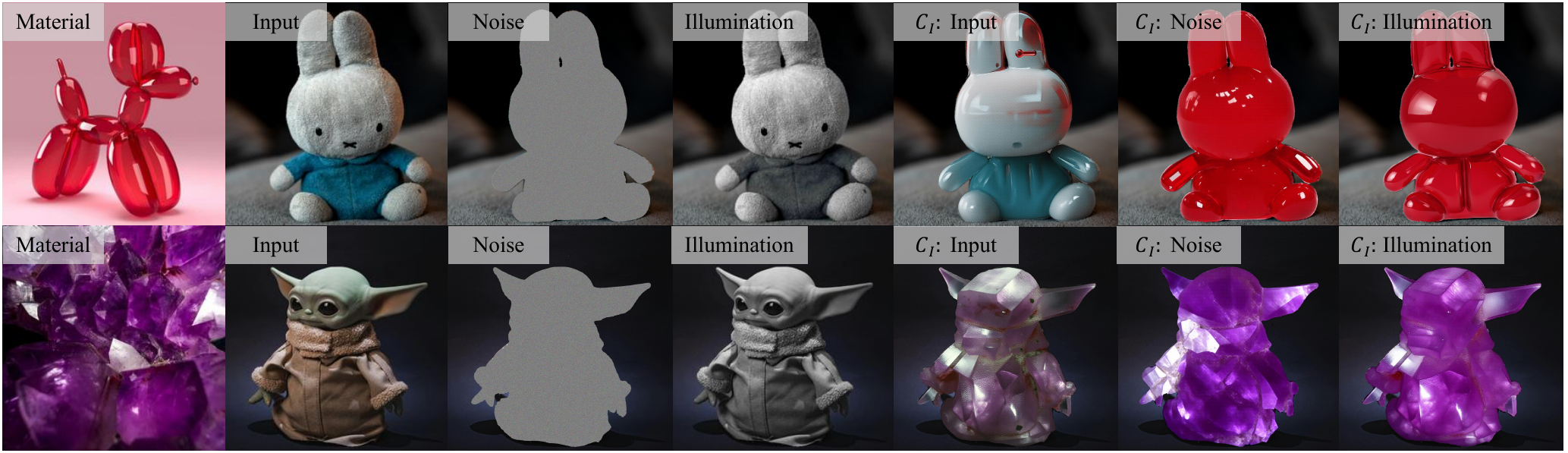}
\vspace{-6mm}
\caption{Regarding the ablation experiment results when $C_I$ is set to different images.
}
\label{fig:ablation_light}
\vspace{-8pt}
\end{figure}
\begin{figure}
\centering
\includegraphics[width= \linewidth]{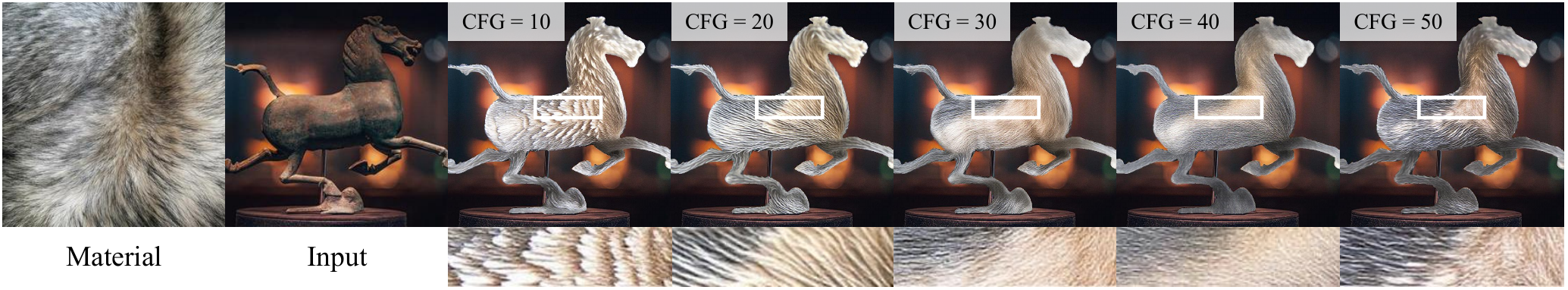}
\vspace{-5mm}
\caption{The visualization results when CFG changes linearly.
}
\label{fig:ablation_cfg}
\vspace{-10pt}
\end{figure}

\subsection{Ablation Studies}
\label{ablation}
We conduct ablation experiments to study the effects of varying 1) material image influence strength, 2) depth image influence strength, 3) optical restoration, and 4) classification guidance (CFG) influence strength. 
As illustrated in Figure~\ref{fig:ablation_strength}, when the Cross-Bias strength factor \(\gamma\) approaches $0$, the material information exerts minimal influence. A satisfactory material transfer effect is achieved as \(\gamma\) increases to $1.8$. However, further increments of \(\gamma\) beyond this value lead to diminishing returns, with little additional improvement observed.
Fig.~\ref{fig:ablation_lora_depth} illustrates the outcomes of controlling depth with different strengths of LoRA influence, where \(w\) takes values in \(\{0.7, 0.8, 0.9, 1.0\}\). The results demonstrate that \(w\) can effectively control the influence of depth information, thus allowing for flexible control over its impact.
Furthermore, to validate the effectiveness of illumination image restoration for input image lighting, we compare the generated results with those using the original image and random noise as inputs. As shown in Fig.~\ref{fig:ablation_light}, the original input image significantly interferes with the color of the production results, while random noise completely disregards the lighting information of the input image, causing the lighting results to lean more towards the material image.
Lastly, we run inferences with CFG strengths of \(\{10, 20, 30, 40, 50\}\). Fig.~\ref{fig:ablation_cfg} shows that increasing CFG strength enhances image detail and fidelity. However, beyond a certain threshold, color distortion and high-frequency noise appear. In our experiments, we consistently use a CFG strength of $30$, which we find to be optimal for our purposes.

\begin{figure}
\centering
\includegraphics[width= \linewidth]{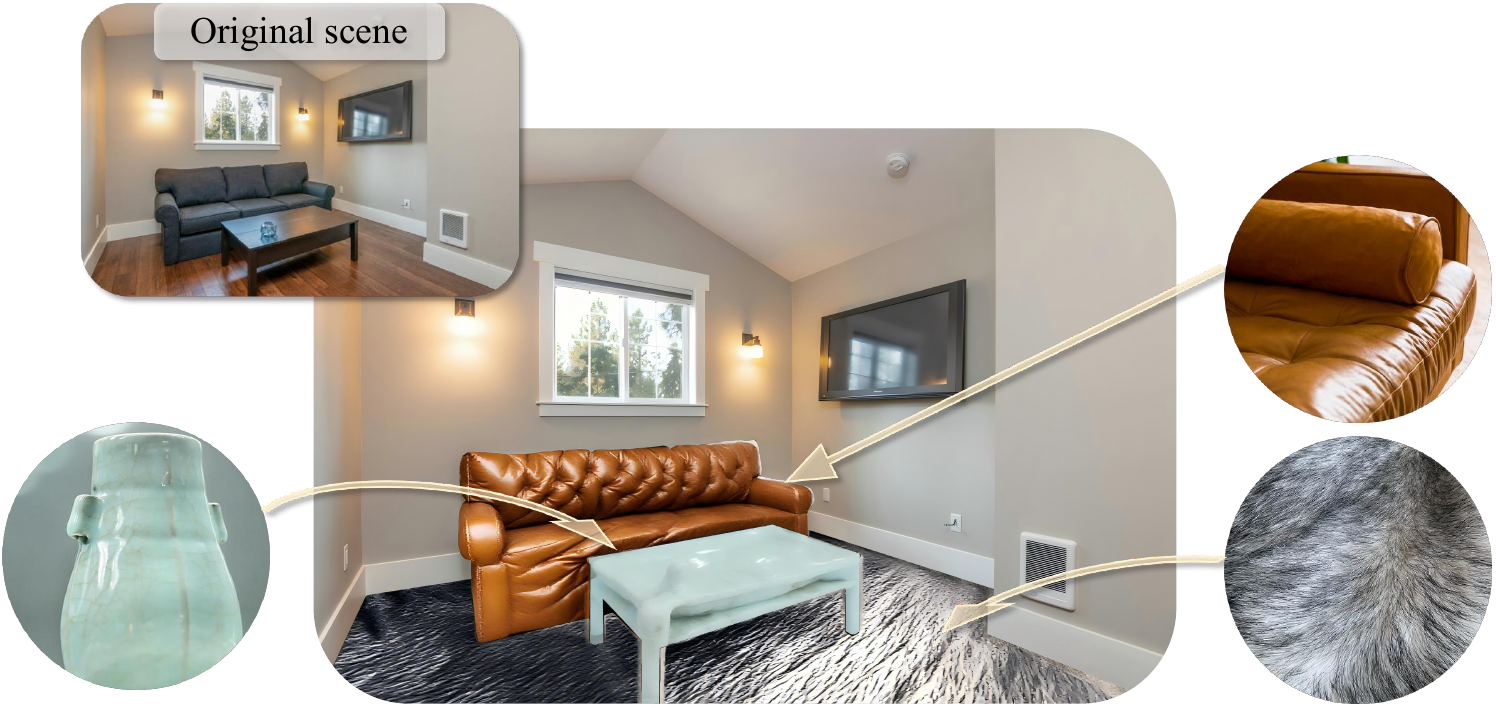}
\vspace{-6mm}
\caption{Multi-object material transfer in the real-world scene.
}
\label{fig:application}
\vspace{-8pt}
\end{figure}
\begin{figure}
\centering
\includegraphics[width= \linewidth]{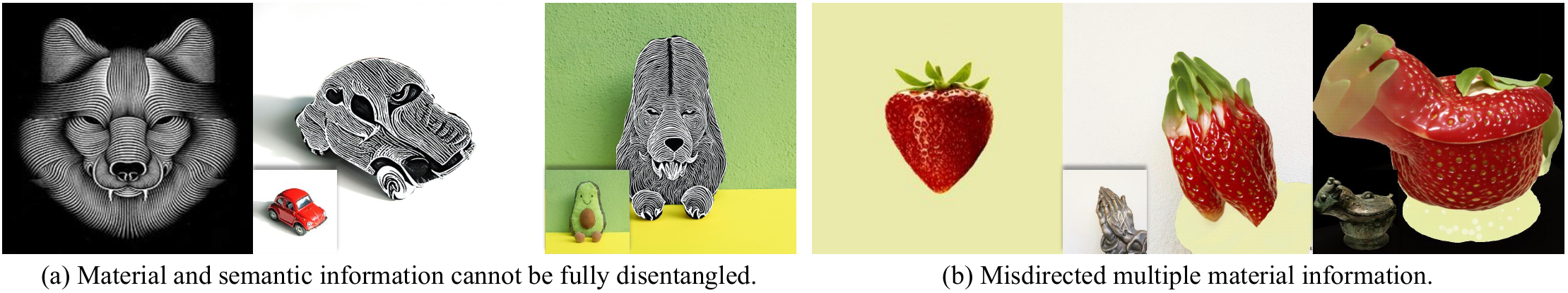}
\vspace{-5mm}
\caption{Limitations. When the semantic information in the material image is too entangled with the texture information, MaTe may struggle to transfer only the texture information.
}
\label{fig:limitation}
\vspace{-10pt}
\end{figure}

\subsection{Applications}
By replacing the foreground extraction with a segmentation module (such as SAM~\cite{ravi2024sam2,zhang2024evfsamearlyvisionlanguagefusion}) to obtain multiple masks, a variety of materials can be applied to multiple objects. Fig.~\ref{fig:application} demonstrates that MaTe can edit multiple objects within a single image, as illustrated by the material editing performed on the sofa, table, and floors in the image, showcasing MaTe's robustness and its applicability to editing complex real-world scenarios.

\subsection{Limitations}
Based on our observations, MaTe currently has two main limitations: 1) If the material is closely entangled with its semantic information, some of the semantic information may also be transferred to the target object. For instance, in Fig.~\ref{fig:limitation} (a), the wolf information from the line art is transferred onto an avocado. 2) Since there are no restrictions on the input material images, it is possible that multiple material information from the materials may be transferred. For example, in Fig.~\ref{fig:limitation} (b), the yellow background of the material image is transferred to the result.
\section{Conclusion}
In this paper, we present MaTe, a streamlined and efficient diffusion model for material transfer that operates without additional training or fine-tuning. By enabling semantic alignment and information mining across illumination, material, and depth images, MaTe achieves material transfer from arbitrary 2D images to target objects while simplifying the inference workflow. Extensive experiments demonstrate that MaTe yields state-of-the-art qualitative and quantitative performance, surpassing existing methods while maintaining a compact and computationally efficient architecture. 
\section{Acknowledgment}
The work is supported in part by the Science and Technology Innovation 2030 – major project brain-inspired learning and game theory for non-cooperative heterogeneous multi-agent systems, China, under Project No. 2021ZD0201404, and the National Science and Technology Council, Taiwan, under Project No. 113-2221-E-006-161-MY3.

{
    \small
    \bibliographystyle{ieeenat_fullname}
    \bibliography{main}
}

\end{document}